\documentclass[10pt,journal,compsoc]{IEEEtran}
\ifCLASSOPTIONcompsoc
  \usepackage[nocompress]{cite}
\else
  \usepackage{cite}
\fi
\usepackage{graphicx}
\usepackage{amsfonts}
\usepackage{multirow}

\ifCLASSINFOpdf
\else
\fi
\usepackage{xcolor}

\usepackage{amsmath}
\definecolor{COLOR}{rgb}{0, 0, 0}
\definecolor{COLOR2}{rgb}{0, 0, 0}
\definecolor{COLOR3}{rgb}{1.0, 0.25, 0.25}
\usepackage{booktabs}

\begin{document}
\title{Node2Seq: Towards Trainable Convolutions in Graph Neural Networks}

\author{Hao~Yuan,
    and Shuiwang~Ji,~\IEEEmembership{Senior~Member,~IEEE}
    % <-this % stops a space
    \thanks{H. Yuan is with Department of Computer Science and Engineering, Texas A\&M University, College Station, TX 77843, USA, e-mail: (hao.yuan@tamu.edu).}% <-this % stops a space
    \thanks{S. Ji is with Department of Computer Science and Engineering, Texas A\&M University, College Station, TX 77843, USA, e-mail: (sji@tamu.edu).}% <-this % stops a space
}


\markboth{Preprint}%
{Yuan \MakeLowercase{\textit{et al.}}: Node2Seq: Towards Trainable Convolutions in Graph Neural Networks}




\IEEEtitleabstractindextext{%
\begin{abstract}
Investigating graph feature learning becomes essentially important with the emergence of graph data in many real-world applications. Several graph neural network approaches are proposed for node feature learning and they generally follow a neighboring information aggregation scheme to learn node features. While great performance has been achieved, the weights learning for different neighboring nodes is still less explored. In this work, we propose a novel graph network layer, known as Node2Seq, to learn node embeddings with explicitly trainable weights for different neighboring nodes. For a target node, our method sorts its neighboring nodes via attention mechanism and then employs 1D convolutional neural networks (CNNs) to enable explicit weights for information aggregation. In addition, we propose to incorporate non-local information for feature learning in an adaptive manner based on the attention scores. Experimental results demonstrate the effectiveness of our proposed Node2Seq layer and show that the proposed adaptively non-local information learning can improve the performance of feature learning.
\end{abstract}

}

\maketitle

\IEEEdisplaynontitleabstractindextext

% \fi
\IEEEpeerreviewmaketitle

\IEEEraisesectionheading{\section{Introduction}\label{sec:introduction}}
\IEEEPARstart{G}{raph}  data are widely existing in different real-world applications, which raises the demand of developing deep learning models for graphs~\cite{wu2020comprehensive,yuan2020explainability,zhang2019graph}. Recently, graph neural networks have achieved great success in many graph-related tasks, such as node classification~\cite{gao2018graph, velivckovic2017graph, liu2020non}, graph classification~\cite{zhang2018end, lee2019self}, link prediction~\cite{kipf2016variational, cai2020link, cai2020line}, and molecular exploration~\cite{wang2020moleculekit}. Several approaches are proposed to investigate different operations for graph neural networks, including node feature learning~\cite{gilmer2017neural, hamilton2017inductive, kipf2016semi}, graph representation learning~\cite{xu2018how}, and graph pooling~\cite{wang2020second,Yuan2020StructPool:, ying2018hierarchical, gao2020topology}. Node feature learning is an important topic in graph neural networks since nodes are the most basic components in the graphs. However, feature learning for graph data is challenging because unlike images and texts, graphs have no locality information. There is no ordering information among different graph nodes and each node can have a variable number of neighboring nodes, which prevents applying traditional convolutional neural networks to graphs. Several approaches are recently proposed to learn node features for graph data, such as graph convolution networks~\cite{kipf2016semi}, graph attention networks~\cite{velivckovic2017graph, gao2019graph,thekumparampil2018attention}, PATCHY-SAN~\cite{niepert2016learning}, and learnable graph convolutional network~\cite{gao2018large}. These feature learning methods generally follow the same high-level pipeline to learn node features that for each node, its new features are obtained by aggregating its neighboring node features. However, most existing approaches cannot learn explicit weights for different neighboring nodes while it is natural that different nodes may have different contributions to their neighbors.

In this work, we propose a novel graph network layer, known as Node2Seq, to enable explicitly trainable weights for different neighboring nodes during the information aggregation procedure. For each node, our Node2Seq builds the ordering information into its neighboring nodes via the attention mechanism and rearrange the feature matrix to follow the ordering information. Then we incorporate 1D CNNs~\cite{lecun1998gradient} to enable explicitly learnable weights for different neighbors.  Next, to address the challenge that different nodes have variable numbers of neighboring nodes, we incorporate the global readout function to learn a global representation for all neighboring nodes. Finally, we also propose to include non-local node information into the information aggregation. Based on the attention scores, the model learns to decide how much non-local information to be included in an adaptive manner. We conduct experiments on several benchmark datasets to evaluate our proposed Node2Seq. Experimental results show that our proposed Node2Seq can outperform comparing baselines significantly. Furthermore, it is also shown that the feature learning results are consistently improved by incorporating our proposed adaptively non-local information learning.

\section{Related Work}
\label{related}
Unlike grid-like data, such as images and texts, graph data contains entity information and cross-entity relationship information but does not have locality information. Generally, a graph can be represented by a feature matrix and an adjacency matrix, which is different from grid-like data. Recently, deep graph neural networks (GNNs) have shown great performance on the graph data. Several GNNs approaches are proposed, such as graph convolution networks (GCNs)~\cite{kipf2016semi}, graph attention networks (GATs)~\cite{velivckovic2017graph}, graph isomorphism networks (GINs)~\cite{xu2018how}, PATCHY-SAN~\cite{niepert2016learning}, and learnable graph convolutional network (LGCN)~\cite{gao2018large}. These GNN methods generally follow a neighborhood information aggregation scheme that for each node, its new features are obtained by aggregating the features of its neighboring nodes and combining them with its own features.
Formally, we represent a graph $G$ as its feature matrix $X \in \mathbb{R}^{n\times d}$ and its adjacency matrix $A \in \{0,1\}^{n\times n}$. We assume that there are $n$ nodes in the graph $G$ and each node has a $d$-dimension feature vector. Then the graph convolution operation is formally defined as
\begin{equation}\label{eq:1}
X' = f(D^{-\frac{1}{2}}\hat{A}D^{-\frac{1}{2}}XW),
\end{equation}
where $X \in \mathbb{R}^{n\times d_{i}}$ and $X' \in \mathbb{R}^{n\times d_{i+1}}$ are the input and output feature matrices of a graph convolution layer. The matrix $D$ denotes the diagonal node degree matrix to normalize $\hat{A}$ and $\hat{A}=A+I$ is the adjacency matrix with self-loops. In addition, $W$ is a learnable matrix to perform linear transformation for features and $f(\cdot)$ is the non-linear function. Obviously, the weights for different neighboring nodes during information aggregation are determined by the node degrees $D^{-\frac{1}{2}}\hat{A}D^{-\frac{1}{2}}$, which are fixed for a given graph. However, different neighboring nodes may contribute differently so that fixed weights may become a limitation for node feature learning. Instead, in the GAT operation, the weights are determined by node similarities and its information aggregation scheme can be expressed as
\begin{equation}\label{eq:2}
x_i' = f(\sum_{j\in\mathcal{N}_i}\alpha_{ij}Wx_j),
\end{equation}
where $x_i'$ is the output feature of node $i$ and $x_j$ is the input feature of node $j$. The neighboring node set of node $i$ is denoted as $\mathcal{N}_i$ and $\alpha_{ij}$ is the attention score between node $i$ and node $j$. In graph attention operation, each node is attended to its neighboring nodes and the weights for different neighboring nodes during information aggregation are the attention scores. Such weights are computed based on the similarities between node features and can be learned indirectly. However, these weights cannot be explicitly learned and it may not be proper to directly use similarities as weights.

None of GCNs and GATs can incorporate explicit weights for different neighboring nodes. To address this limitation, the PATCHY-SAN~\cite{niepert2016learning} and LGCN~\cite{gao2018large} propose to employ traditional CNNs to enable explicitly learnable weights. The PATCHY-SAN first orders different nodes using graph kernels and selects a fixed length of node sequence. Then for each node, PATCHY-SAN selects a fixed size of neighboring nodes such that different nodes are ordered and each node has the same receptive field. Finally, the PATCHY-SAN applies transitional CNNs to learn explicit weights for feature learning. However, the transformation is designed as a data preprocessing step and the model cannot be trained in an end-to-end manner. In addition, by selecting a fixed number of nodes and a fixed number of neighbors, the PATCHY-SAN may ignore important node information, which can affect the feature learning results. In addition, the LGCN proposes an end-to-end framework to incorporate 1D CNNs in node feature learning.  For each node, the LGCN selects the top-$k$ features along each channel dimension from its neighboring nodes, and the selection is based on the numerical values of features. Then each node has the same size of neighboring feature representations and LGCN applies 1D CNNs to learn new features for the central target nodes. However, the feature-wise selection does not consider the node-level information and affects the consistency between features and nodes. Furthermore, selecting top features based on numerical values may ignore important information since a feature can be important but with a small numerical value. Different from existing works, our proposed method learns to order neighboring nodes via attention mechanism so that it can be trained in an end-to-end manner. In addition, our method does not need to select a portion of neighboring nodes and can avoid the information loss during selections. Furthermore, our method can adaptively incorporate non-local information and the model automatically learns how the non-local information is used.

\section{The Proposed Methods}
\label{method}
In this work, we propose a novel graph neural network layer, known as the Node2Seq layer. It enables us to learn node embeddings by aggregating neighborhood information with explicitly learnable weights. It employs the attention mechanism to order neighboring nodes and traditional convolutional networks to assign different weights to them. In addition, we propose to incorporate non-local information by adaptively ranking and selecting nodes with the attention scores.

\subsection{The Node2Seq Layer}
Convolutional Neural Networks (CNNs) are shown to be effective for featuring learning on grid-like data, such as images and texts~\cite{ji20123d, kim2014convolutional, he2016deep}. The information in neighborhood  is aggregated by explicitly learnable weights. Since graphs have no locality information, CNNs cannot be directly applied to graph data. Existing approaches, such as GCNs~\cite{kipf2016semi} and GATs~\cite{velivckovic2017graph}, are proposed to aggregate neighborhood information for graph data. However, there are no explicitly trainable weights, as the weights in CNNs, to measure the contributions of different nodes during the aggregation procedure.

\begin{figure*}[!ht]
    \centering
    \includegraphics[width=2\columnwidth]{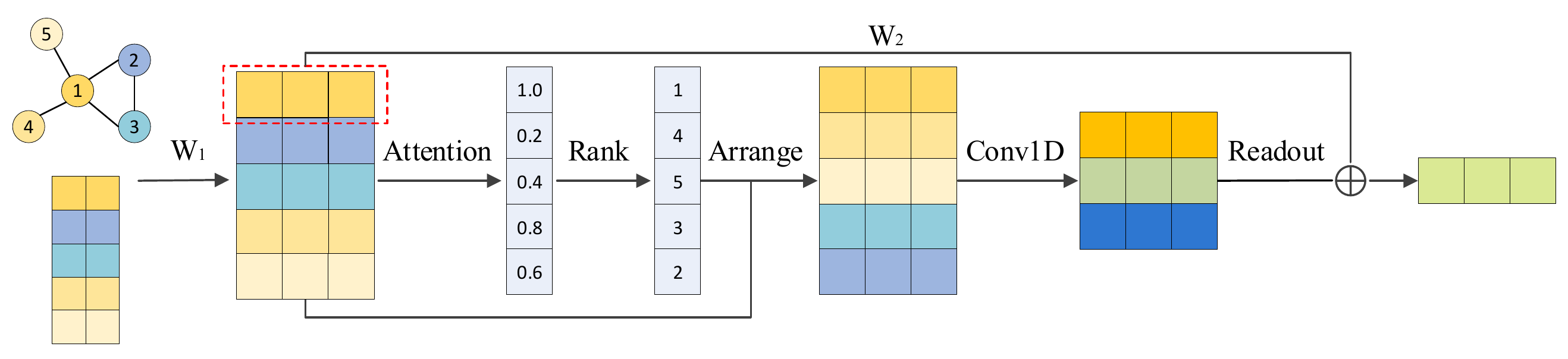}
    \caption{An illustration of our proposed Node2Seq for node feature learning. Given the input graph with 5 nodes where each node has a two-dimensional feature vector and the colors represent different features. The goal is to learn new features for node 1, based on the input feature matrix.         We first perform a linear transformation $W_1$ on the input and the channel dimension is increased from 2 to 3. Then the attention scores are computed by attending node 1 to other nodes, including node 1 itself. We rank the attention scores in the descending order and obtained the corresponding indices. Clearly, based on the similarity to node 1, the indexing order is node 1, node 4, node 5, node 3, and node 2.
        Note that for simplicity, the example does not include 2-hop neighbors for node 1 and such nodes are ignored if existing. Next, the node feature matrix is rearranged to follow the indexing order. Then we perform traditional 1D CNNs on the ordered features obtain a $3\times 3$ output. Here the kernel size is set to 3, stride size is set to 1, and no padding is added. Finally, we obtain a $1\times 3$ representation via the Readout function and combine it with the features of node 1 to obtain the new features.}
    \label{fig:pip}
\end{figure*}
In this section, we introduce our proposed Node2Seq layer, which incorporates 1D CNNs to perform neighborhood aggregation. Note that there are two main challenges to apply traditional CNNs to graph data; those are, there is no ordering information among different nodes and the numbers of neighboring nodes vary for different nodes. To address these challenges, our proposed Node2Seq layer first learns to rank neighboring nodes via attention mechanism~\cite{vaswani2017attention} that for a given node $i$, its neighboring nodes are attended to this node and the attention scores are computed. The attention score between two nodes indicates how similar these two nodes are. Then based on the attention scores, the neighboring nodes of node $i$ are sorted and hence the ordering information becomes explicit. Next, 1D CNNs are employed to learn different weights for aggregating different node information. Finally, we combine the aggregated output features and the features of node $i$ to obtain the new features for node $i$.
Formally, given an input graph $G$ with $n$ nodes, it is represented by its adjacency matrix $A \in \{0,1\}^{n\times n}$ and its feature matrix $X \in \mathbb{R}^{n\times c}$. For the node $i$, it has a $c$-dimensional feature vector, denoted as $x_i$. Its edge connections with other nodes are indicated by the $i$th row of $A$, denoted as $a_i$. We illustrate our method by showing how to learn new features for node $i$ by aggregating its neighbor nodes. The forward propagation rules of our Node2Seq layer are mathematically written as
\begin{align}
\bar{X} & = XW_1, & \in \mathbb{R}^{n\times c'} \label{eq1}\\
%\bar{X}_m & = \mbox{Extract}(\bar{X}, \bar{A}, i), & \in \mathbb{R}^{m\times c'}\\
s_i & = \bar{x}_i\bar{X}^T, & \in \mathbb{R}^{1\times n} \label{eq2}\\
idx & = \mbox{Ranking}(s_i, \bar{A}), & \in \mathbb{R}^{1\times k_i} \label{eq3}\\
X_s &= \mbox{Arrange}(\bar{X}, idx), & \in \mathbb{R}^{k_i \times c'} \label{eq4}\\
F_i &= \mbox{Conv1D}(X_s),  & \in \mathbb{R}^{o_i \times c'} \label{eq5}\\
out &= \mbox{Readout}(F_i),  & \in \mathbb{R}^{1 \times c'} \label{eq6}\\
\widehat{x_i}& = \bar{x_i}W_2 + out, & \in \mathbb{R}^{1 \times c'} \label{eq7}
\end{align}
where $W_1\in \mathbb{R}^{c\times c'}$ and $W_2\in \mathbb{R}^{c\times c'}$ are trainable weight matrices. The operation $\mbox{Ranking}(\cdot,\cdot)$ returns the indices that can sort the attentions scores considering the graph connectivity information. Note that self-loops are added to the adjacency matrix as $\bar{A}= A + I$ and $k_i$ denotes the number of neighboring nodes for node $i$. The $\mbox{Arrange}(\cdot,\cdot)$ operation extracts and rearranges the rows of feature matrix to follow the order of indices $idx$.  In addition, $\mbox{Readout}(\cdot)$ reduces the spatial sizes of feature matrix from $o_i \times c'$ to $1 \times c'$.

To address the first challenge that there is no ordering information among different nodes, we employ the attention mechanism to build such ordering information into neighboring nodes. We first perform a linear transformation in Eq. (\ref{eq1}) via learnable matrix $W_1$ to learn the new feature matrix $\bar{X}$ for the input feature matrix $X$. Then, as shown in Eq. (\ref{eq2}), the attention scores $s_i$ are computed by attending node $i$ to other nodes, which represents the similarities between node $i$ and the other nodes. In our attention mechanism, the node features of node $i$ serve as the query while the keys are node features of other nodes.
Next, based on the attention scores, we can rank the neighboring nodes and obtain the corresponding indices in Eq. (\ref{eq3}). We only consider the neighboring nodes connected with node $i$ for node ranking while ignoring the rest nodes. Assuming that node $i$ has  $k_i$ 1-hop neighboring nodes indicated in $\bar{A}$,
the ranking operation $\mbox{Ranking}(\cdot,\cdot)$ only returns $k_i$  indices such that the corresponding nodes are sorted by the attentions scores in the descending order.  Note that in the implementation, we can compute the attention scores for $k_i$ 1-hop neighboring nodes and then rank them to obtain the indices.
With such indexing information, we build the ordering information for neighboring nodes and rearrange the feature matrix. The $\mbox{Arrange}(\cdot,\cdot)$ in Eq. (\ref{eq4}) operation extracts the corresponding $k_i$ rows of $\bar{X}$ and arrange them to follow the $idx$ order.
For node $i$, its $k_i$ neighboring nodes are extracted and their features explicitly ordered, denoted as $X_s$. In this way, the first challenge is addressed and traditional CNNs can be applied. We apply 1D CNNs in Eq. (\ref{eq5}) to the ordered features $X_s$ to learn explicit weights for aggregating information from different neighbors.

With 1D CNNs, we can obtain the aggregated features for neighboring nodes, denoted as $F_i$. Since the neighboring node number $k_i$ varies for different node $i$, the spatial size of the output, denoted as $o_i$, is also different for different $i$. Then such output $F_i$ cannot be directly incorporated, which corresponds to the second aforementioned challenge. Hence, in Eq. (\ref{eq6}), we perform a readout function on the output feature $F_i$. The $\mbox{Readout}(\cdot)$ is employed to combine $o_i$ feature vectors and obtain a single vector as the global representation for all neighboring nodes. It reduces the spatial sizes of $F_i$ from $o_i \times c'$ to $1 \times c'$ and for all different nodes, the $F_i$ are reduced to the same dimensions regardless of its neighboring number $k_i$. Note that there are several choices for the  $\mbox{Readout}(\cdot)$ operation, such as global max pooling, global average pooling, and global sum pooling. Finally, we combine the neighboring information and the embeddings of node $i$ to obtain the new embeddings of node $i$ in Eq. (\ref{eq7}). To this end, we perform another linear transformation $W_2$ on the features of node $i$ and combine it with the global representation for all neighboring nodes $out$ via a simple summation. Overall, our proposed Node2Seq layer learns new embeddings $\widehat{x_i}$ for node $i$ based on its original embeddings $x_i$ and its neighboring information. The neighboring information is aggregated with explicitly learnable weights via traditional 1D CNNs.
\begin{figure*}[!ht]
    \centering
    \includegraphics[width=2\columnwidth]{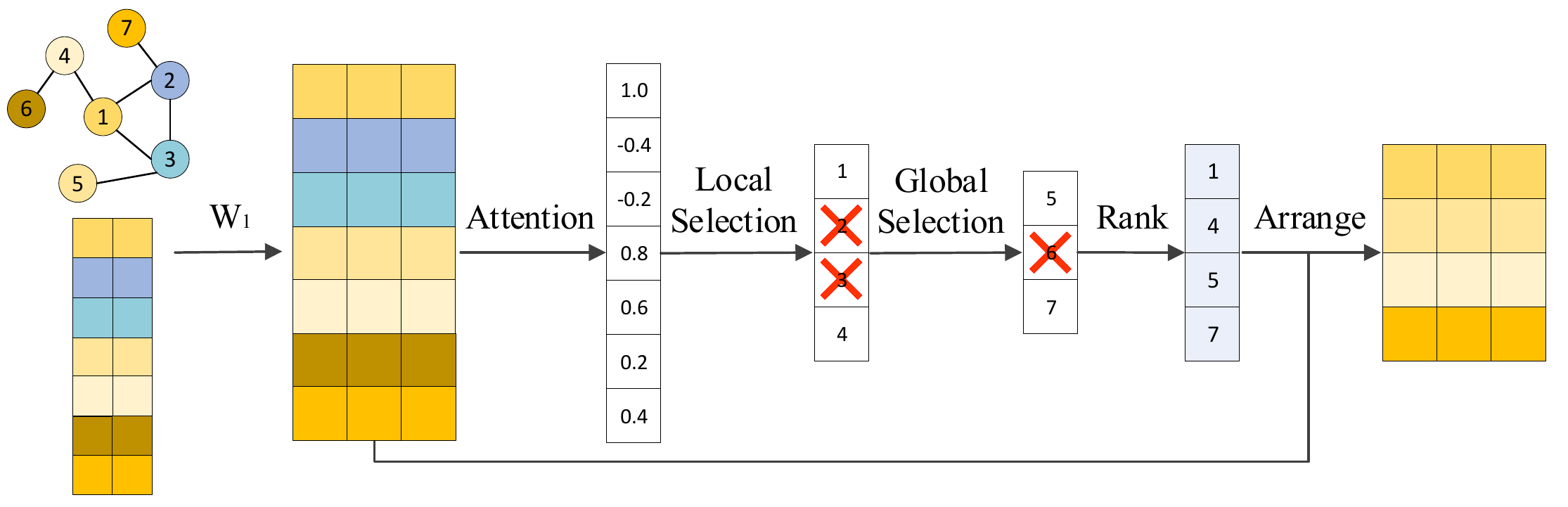}
    \caption{An illustration of our proposed non-local information learning. Given the input graph with 7 nodes where different colors represent different features. We show the steps to learn non-local information for node 1. With the linear transformation $W_1$, the channel dimension is increased from 2 to 3 and the feature matrix is $7\times 3$. In this example, we set $\ell=2$ that information within 2-hops is considered. The attention scores are computed by attending node 1 to other nodes, including node 1 itself. Based on the attention scores, we first select local nodes for it. For node 1, there are $k_i= 4$ 1-hop nodes (nodes 1, 2, 3, 4) and $m_i=2$ (nodes 1, 4) are selected when we set $\beta=0$. The red crosses denote the corresponding nodes are not selected. Next, we include $k_i-m_i=2$ non-local nodes from $\bar{A}^{2}$ neighboring (node 5, 7). Finally, those selected nodes are used to build the ordered feature matrix.}
    \label{fig:non-local}
\end{figure*}

Our proposed Node2Seq layer also follows the neighborhood aggregation strategy as other GNNs. Intuitively, each sliding window in 1D CNNs learns the aggregation of a portion of neighboring nodes, and then the global representation of all neighboring nodes is obtained to update the embeddings of the target node. The use of 1D CNNs not only learns weights for different nodes but also captures relationships across different feature channels. The general pipeline of our proposed Node2Seq is shown in Figure~\ref{fig:pip}, where we illustrate how to learn new features for the target node 1 and aggregate information from its 4 neighboring nodes.

\subsection{Non-Local Information Learning}\label{non-local}
As mentioned above, our proposed Node2Seq incorporates the 1-hop connectivity information $\bar{A}$ to guide the aggregation strategy. For a target node, only its own information and 1-hop neighboring information are used to learn new features in Eq. (\ref{eq3}). Recently, existing work~\cite{wang2018non} shows that learning global information is important and can better capture the relationships among different input regions for image and NLP tasks. Similarly, for graph data, it is not only necessary to capture the relationships between local neighboring nodes, but also important to model relationships between non-local nodes. Hence, based on our proposed Node2Seq, we further improve it to capture such non-local information and include non-local neighboring nodes in the aggregation procedure. Incorporating more nodes can aggregate more information for feature learning; however, it significantly increases the computational cost and may lead to the over-fitting problem. Then it is challenging to determine how much non-local information should be used. Intuitively, for certain nodes, the local nodes are enough for the feature learning; however, the non-local information can be important for other nodes. Hence, we propose to adaptively incorporate the non-local information.

We employ the attention scores to adaptively incorporate the non-local information. Specifically, we improve the index ranking step in Eq. (\ref{eq3}) that we adaptively include non-local nodes and ignore some local nodes. Formally, let $\ell$ define the number of neighboring hops to be considered, and $A^{\ell}\in \{0,1\}^{n\times n}$ represent the $\ell$-hop connectivity of the graph $G$.  By adding the self-loops, the connectivity becomes $\bar{A}^{\ell}= A^{\ell} + I$ and $\bar{a}_{i,j}^{\ell}=1$ indicates node $i$ and node $j$ are reachable in $G$ within $\ell$ hops.
Then, for node $i$, which has $k_i$ 1-hop neighboring nodes, its index ranking becomes three steps: local nodes selection, non-local node selection, and index ranking. First, for any 1-hop neighboring node in $\bar{A}$, it is selected if its attention score $s_{ij}>\beta$ where $\beta$ is a predefined threshold. Let $m_i \leq k_i$ denotes the number of selected 1-hop local nodes and their indices are ranked by the attention scores. Next, if $m_i$ is equal to $k_i$, we believe using local nodes is enough to learn new features for node $i$. Otherwise, we include $k_i-m_i$ non-local nodes from $\bar{A}^{\ell}$ and store their indices. Finally, $k_i$ nodes are selected and we combine the aforementioned two sets of indices as the ranking results in Eq. (\ref{eq3}).
Note that no matter how much non-local information is incorporated, only $k_i$ nodes will be used for information aggregation. Once a non-local node is selected, a local node is ignored.
Since the attention scores are learned by the model, such a trade-off is automatically learned by the model, and the non-local information is incorporated in an adaptive way.
When $\ell$ is set to 1, only node information within 1-hop is considered. If $\ell$ is large enough, the whole global node information can be incorporated.
We illustrate our non-local information learning in Figure~\ref{fig:non-local}, where we show how to adaptively use 2-hop neighbors.

\begin{table*}[!ht]
    \caption{Statistics and properties of  benchmark datasets.}
    \label{table:data}
    \centering
    \begin{tabular}{@{}lccccccccc@{}}
        \toprule
        \multicolumn{1}{l}{ }       &    \multicolumn{9}{c}{\textbf{Dataset}}          \\
        \cmidrule(r){2-10}
        & \textsc{Cora}     & \textsc{Citeseer}& {\textsc{Amazon Photo}} & {\textsc{Chameleon}}  &   {\textsc{Squirrel}} & {\textsc{Actor}} & \textsc{Cornell} & \textsc{Texas} & \textsc{Wisconsin} \\
        \midrule

        \# of Nodes   & 2708  & 3327 &7487 &2277 & 5201& 7600 & 183 & 183 & 251    \\
        \# of Edges      & 5429  &4732  &119043 &36101   & 217073&33544  & 295 & 309 & 499  \\
        \# of Features     & 1433  & 3703 &745 &2325  & 2089& 931 & 1703 & 1703 & 1703    \\
        \# of Classes     & 7  & 6 &8 &5   & 5& 5 & 5& 5   & 5      \\
        \bottomrule
    \end{tabular}
\end{table*}
\begin{table*}[ht!]
    \caption{Node classification results for three datasets. Both fixed and random split are evaluated.}
    \label{table:res}
    \centering
    \begin{tabular}{@{}lccccc@{}}
        \toprule
        \multicolumn{1}{l}{\textbf{Method}}       &    \multicolumn{5}{c}{\textbf{Dataset}}          \\
        \cmidrule(r){2-6}
        & \textsc{Cora-F}     & \textsc{Cora-R}& {\textsc{Citeseer-F}} & {\textsc{Citeseer-R}} &  {\textsc{Amazon-R}}  \\
        \midrule

        \textsc{Cheby}  & 80.7$\pm$ 1.1  & 76.5$\pm$ 1.6 &69.9$\pm$ 1.0 &67.2$\pm$ 1.6 & 88.0$\pm$ 1.9    \\
        \textsc{GCNs}      & 81.6$\pm$ 0.8  & 79.2$\pm$ 1.9 &70.3$\pm$ 0.9 &67.9$\pm$ 1.7 & 87.6$\pm$ 2.1    \\
        \textsc{GATs}     & 82.9$\pm$ 0.5  & 81.1$\pm$ 1.2 &70.7$\pm$ 0.6 &69.1$\pm$ 1.6 & 89.4$\pm$ 1.8    \\
        \textsc{SGC}      & 81.2$\pm$ 1.6  & 80.7$\pm$ 1.8 &71.0$\pm$ 0.5 &68.3$\pm$ 1.4 & 89.5$\pm$ 1.8    \\
        \textsc{LGCN}      & 82.7$\pm$ 0.7  & 80.5$\pm$ 1.5 &71.2$\pm$ 0.6 &68.6$\pm$ 1.7 & 88.9$\pm$ 1.9    \\
        \midrule
        \textsc{Node2Seq} & \bf{84.0$\pm$ 0.6} & \bf{82.0$\pm$ 1.2} &\bf{72.4$\pm$ 1.1} & \bf{70.8$\pm$ 1.7} & \bf{90.1$\pm$ 1.1}   \\
        \bottomrule
    \end{tabular}
\end{table*}

\section{Experimental Studies}
\label{exp}
\subsection{Datasets and Experimental Setup}
To demonstrate the effectiveness of our proposed Node2Seq, we evaluate our method on several datasets, including citation network datasets~\cite{sen2008collective}, WebKB datasets, Amazon co-purchase datasets~\cite{mcauley2015image}, Actor co-occurrence datasets~\cite{tang2009social}, and Wikipedia network datasets~\cite{rozemberczki2019multi}. The citation networks are dataset Cora and Citeseer, where graph nodes represent academic papers, and graph edges denote the citation relationships between different papers. The WebKB data contain webpage dataset Cornell, Texas, and Wisconsin, which collects the webpages of computer science departments from different schools. In these datasets, graph nodes are web pages while the edges correspond to the hyperlinks among different web pages. For the
Amazon co-purchase data, we use the Amazon Photo dataset, where nodes mean different products and edges indicate that the connected products are frequently bought
together. The Actor co-occurrence dataset is known as Actor, which contains nodes representing actors and edges representing if two actors are on the Wikipedia page. For the Wikipedia network data, we use two datasets known as Chameleon and Squirrel, which contains graph nodes representing Wikipedia pages and edges representing the mutual links between pages.  Note that for all datasets, all graphs are treated as undirected. We report the statistics and properties of nine benchmark datasets in Supplementary Table~\ref{table:data}. We implement our Node2Seq models using Pytorch~\cite{paszke2017automatic} with the PyTorch geometric framework~\cite{Fey/Lenssen/2019} and conduct experiments on one GeForce GTX 1080 Ti GPU.
The model is trained using  Stochastic gradient descent (SGD) with the ADAM optimizer~\cite{kingma2014adam}.

We compare our proposed Node2Seq with several feature learning methods; those are, GCNs~\cite{kipf2016semi}, GATs~\cite{velivckovic2017graph}, Cheby~\cite{defferrard2016convolutional}, SGC~\cite{wu2019simplifying}, and LGCN~\cite{gao2018large}. The GCNs learn node features by aggregating 1-hop neighboring nodes and the weights are determined by the node degrees.  The GATs can perform masked attention to incorporate both local and non-local information based on the adjacency matrix and the inexplicit weights of different aggregated nodes are determined by node similarities. The Cheby method proposes a spectral graph-theoretical formulation of CNNs on graphs and learns fast localized spectral filters. The SGC is a simplified version of GCNs which produces a linear model by removing nonlinearities and collapsing weight matrices between consecutive layers. In addition, the LGCN first selects the largest $k$ features along different channels and then applies 1D CNNs on them. To evaluate our proposed Node2Seq layer, we build our model based on the GCN networks. Specifically, we combine 1 layer of our Node2Seq with 2-layer GCN networks. In addition, we incorporate skip connections~\cite{xu2018representation, ronneberger2015u} to facilitate feature learning that the outputs of the first two layers are combined via summation or concatenation. We also employ dropouts in our model to avoid the over-fitting problem. For the LGCN method, we directly use the code released by their authors. For other comparing methods, we use the implementations in PyTorch-geometric~\cite{Fey/Lenssen/2019}. For other comparing methods, we follow the benchmark settings in related studies~\cite{Fey/Lenssen/2019, gao2018large} that we use the two-layer network for GCNs, GATs, and Cheby, three-layer network for LGCN, and one-layer network for SGC.

\subsection{Node Classification Results}

We evaluate our proposed method on nine node classification datasets and compare it with several state-of-the-art feature learning methods. The results are reported in Table~\ref{table:res} and Table~\ref{table:res2} where the best results are shown in bold. We evaluate our method with two different settings. For the results in Table~\ref{table:res}, the training sets are significantly smaller than the validation sets and test sets. For datasets Cora and Citeseer, we use 20 labeled nodes per class as the training set, 500 nodes as the validation set, and the rest nodes as the test set. In addition, we consider both fixed train/val/test split following~\cite{kipf2016semi} and random train/val/test split following~\cite{shchur2018pitfalls}. For the dataset Amazon Photo, we use 20 nodes per class as the training set, 30 nodes per class as the validation set, and the rest nodes as the test set. In Table~\ref{table:res}, we denote the fix split case as ``\textsc{Dataset-F}'' and the random split case as ``\textsc{Dataset-R}''. The reported results are obtained by averaging the results of 20 runs.  Clearly, in Table~\ref{table:res}, our methods can significantly outperform other baselines for both the fixed data split and the random data split. Our method outperforms the second-best performance by an average of 1.1\% in these 5 datasets.

\begin{table*}[ht!]
    \caption{Node classification results for six datasets.}
    \label{table:res2}
    \centering
    \begin{tabular}{@{}lcccccc@{}}
        \toprule
        \multicolumn{1}{l}{\textbf{Method}}       &    \multicolumn{6}{c}{\textbf{Dataset}}          \\
        \cmidrule(r){2-7}
        & \textsc{Chameleon}& {\textsc{Squirrel}} & {\textsc{Actor}} &  {\textsc{Cornell}} &  {\textsc{Texas}}& \textsc{Wisconsin} \\
        \midrule
        \textsc{GCNs}      & 68.1$\pm$ 1.6  & 54.1$\pm$ 1.7 &30.6$\pm$ 1.0 &57.2$\pm$ 7.3 & 61.2$\pm$ 5.6  & \bf{60.7$\pm$ 5.5}  \\
        \textsc{GATs}     & 66.3$\pm$ 2.8  & 50.8$\pm$ 2.1 &29.6$\pm$ 0.8 &55.1$\pm$ 5.9 & 58.1$\pm$ 5.1   & 58.7$\pm$ 7.7 \\
        \textsc{SGC}      & 65.0$\pm$ 2.1  & 45.5$\pm$ 1.8 &30.2$\pm$ 1.3 &55.8$\pm$ 5.3 & 55.1$\pm$ 5.4  & 53.7$\pm$ 6.3  \\
        \midrule
        \textsc{Node2Seq} & \bf{69.4$\pm$ 1.6} & \bf{58.8$\pm$ 1.4} &\bf{31.4$\pm$ 1.0} & \bf{58.7$\pm$ 6.8} & \bf{63.7$\pm$ 6.1} & 60.3$\pm$ 7.0  \\
        \bottomrule
    \end{tabular}
\end{table*}
\begin{table*}[ht!]
    \caption{ Node classification results for the model with only local information and the model incorporating non-local information.}
    \label{table:res3}
    \centering
    \begin{tabular}{@{}lcccccc@{}}
        \toprule
        \multicolumn{1}{l}{\textbf{Method}}       &    \multicolumn{6}{c}{\textbf{Dataset}}          \\
        \cmidrule(r){2-7}
        & \textsc{Cora-F} & {\textsc{Citeseer-F}} &  {\textsc{Citeseer-R}} &  {\textsc{Chameleon}}& \textsc{Squirrel} & {\textsc{Texas}}\\
        \midrule
        \textsc{Node2Seq\_l}      & 81.8$\pm$ 1.3  & 70.7$\pm$ 1.6 & 69.8$\pm$ 1.4 &68.5$\pm$ 2.8 & 57.0$\pm$ 1.3 & 61.5$\pm$ 6.3  \\
        \textsc{Node2Seq\_g} & \bf{84.0$\pm$ 0.6} & \bf{72.4$\pm$ 1.1} &\bf{70.8$\pm$ 1.7} & \bf{69.4$\pm$ 1.6} & \bf{58.8$\pm$ 1.4} & \bf{63.7$\pm$ 6.1}  \\
        \bottomrule
    \end{tabular}
\end{table*}
In addition, we evaluate our method for another setting that more nodes are used in the training than testing. For all datasets in Table~\ref{table:res2}, we randomly split the datasets that we use 60\% nodes per class as the training set, 20\% nodes per class as the validation set, and the rest 20\% nodes as the test set. Similarly, the reported results are obtained by averaging the results of 20 runs. Obviously, our method achieves the best performance on five of six datasets and significantly outperforms all comparing methods. Note that our method outperforms the second-best performance by an average of 1.3\% in these 5 datasets. Considering that the datasets Cornell, Texas, and Wisconsin are relatively small and all methods are not performing stably, the results indicate our Node2Seq perform competitively and even better compared with other methods. For the dataset Wisconsin, the GCNs only slightly outperform our proposed Node2Seq by 0.4\%.  Overall, our method shows promising performance under two different settings and the results demonstrate its effectiveness.

\subsection{Effects of Non-Local Information}

In Section~\ref{non-local}, we propose to adaptively incorporate the non-local neighboring information. We conduct experiments to show  the effectiveness of our proposed adaptively non-local information learning.  Specifically, we compare our Node2Seq with only local information, denoted as \textsc{Node2Seq\_l}, with our Node2Seq with global information, denoted as \textsc{Node2Seq\_g}. Both of them are evaluated under the same model framework and we report the averaged results of 20 runs in the Table~\ref{table:res3}.  Obviously, with non-local information, our \textsc{Node2Seq\_g} can outperform the \textsc{Node2Seq\_l} significantly and consistently. The results show that our proposed adaptively non-local information learning is useful and can improve feature learning results.

\subsection{Ablation Studies}
\begin{table*}[ht!]
    \caption{Comparison between our model and the model replacing our Node2Seq with GCN.}
    \label{table:res4}
    \centering
    \begin{tabular}{@{}lcccccc@{}}
        \toprule
        \multicolumn{1}{l}{\textbf{Method}}       &    \multicolumn{6}{c}{\textbf{Dataset}}          \\
        \cmidrule(r){2-7}
        & \textsc{Cora-F} & {\textsc{Cora-R}}& {\textsc{Citeseer-F}} &  {\textsc{Citeseer-R}} &  {\textsc{Chameleon}}& \textsc{Squirrel} \\
        \midrule
        \textsc{Gcns}      & 81.6$\pm$ 0.8  & 79.2$\pm$ 1.9 &70.3$\pm$ 0.9 &67.9$\pm$ 1.7 & 68.1$\pm$ 1.6 & 54.1$\pm$ 1.7  \\
        \textsc{Gcns*}      & 80.7$\pm$ 1.5  & 79.1$\pm$ 1.7 & 69.2$\pm$ 1.8 &69.1$\pm$ 1.7 & 68.6$\pm$ 2.2 & 56.5$\pm$ 1.4  \\
        \textsc{Ours} & \bf{84.0$\pm$ 0.6} & \bf{82.0$\pm$ 1.2} &\bf{72.4$\pm$ 1.1} & \bf{70.8$\pm$ 1.7} & \bf{69.4$\pm$ 1.6} & \bf{58.8$\pm$ 1.4}  \\
        \bottomrule
    \end{tabular}
\end{table*}
We further perform the ablation study to evaluate our proposed Node2Seq. As mentioned above, our model is a 3-layer network built by combining 1-layer Node2Seq and 2-layer GCNs, which also incorporates skip-connections. To show that the performance gain is mainly obtained by our proposed Node2Seq instead of the network designing, we compare our method with the one replacing our Node2Seq layer by one GCN layer while keeping other settings the same, denoted as \textsc{Gcns*}. The results are reported in Table~\ref{table:res4} and they are obtained by averaging the results of 20 runs. We also show the results of 2-layer GCNs, denoted as \textsc{Gcns}.  By comparing our method and \textsc{Gcns*}, we can conclude that our proposed Node2Seq can improve the performance of feature learning. In addition, by comparing \textsc{Gcns*} and \textsc{Gcns}, we find that adding one GCN layer only leads to incremental improvements on three out of six datasets and performs even worse than the 2-layer GCNs on the other three datasets. Such observations further demonstrate the effectiveness of our proposed method.
\subsection{Analysis of Convolutional Kernels}
\begin{table}[ht!]
    \caption{Node classification results for different kernel sizes.}
    \label{table:res5}
    \centering
    \begin{tabular}{@{}lcccc@{}}
        \toprule
        \multicolumn{1}{l}{\textbf{Dataset}}       &    \multicolumn{4}{c}{\textbf{Kernel Size}}          \\
        \cmidrule(r){2-5}
        & $k=3$ & $k=5$& $k=8$ &  $k=10$ \\
        \midrule
        \textsc{Cora}      & 83.5$\pm$ 0.7  & 83.7$\pm$ 0.4 & 83.8$\pm$ 0.7 &\bf{84.0$\pm$ 0.6}  \\
        \textsc{Squirrel} & \bf{58.8$\pm$ 1.4} & 58.1$\pm$ 1.8 & 57.8$\pm$ 1.7 &  57.6$\pm$ 1.5    \\
        \bottomrule
    \end{tabular}
\end{table}
In our Node2Seq layer, we incorporate the 1D CNNs to enable explicit weights for different neighbors. We perform experiments to evaluate how different kernel sizes $k$ affect the performance of feature learning. The results are reported in Table~\ref{table:res5} and all results are obtained by averaging the results of 20 runs. We find the selection of kernel size highly depends on the dataset at hand. For the dataset Cora, the node classification results are generally increasing with the increase of kernel size. For the dataset Squirrel, the best performance is obtained for $k=3$ and the performance is decreasing with the increase of $k$. In addition, the performance remains competitive and can outperform other comparing methods for different  $k$ values. It further indicates the advantage of our proposed Node2Seq layer in node feature learning.

\section{Conclusions}
\label{con}
Learning node features is important and several deep learning methods are proposed to aggregate neighboring node information. However, methods like GCNs and GATs cannot learn explicit weights for different neighboring nodes. Other methods, such as PATCHY-SAN and LGCN, suffers from important information loss when ignoring nodes and features. In this work, we propose a novel graph neural network layer, known as Node2Seq, to enable explicitly learnable weights for the neighboring information aggregation. We employ the attention mechanism to order neighboring nodes and employ 1D CNNs to learn explicit weights. In addition, we incorporate the readout function to obtain overall representations for all neighboring nodes so that no node information is ignored. Furthermore, we propose to incorporate non-local information in an adaptive manner that the model learns to decide whether the non-local information should be included and how much non-local information is used based on the attention scores. We conduct experiments to demonstrate the effectiveness of our proposed Node2Seq. It is shown that our Node2Seq can improve feature learning results. In addition, experimental results also indicate that the non-local information is important and useful.

\ifCLASSOPTIONcompsoc
  \section*{Acknowledgments}
\else
  \section*{Acknowledgment}
\fi

This work was supported in part by National Science Foundation grant IIS-1955189.

\ifCLASSOPTIONcaptionsoff
  \newpage
\fi



%\bibliographystyle{IEEEtran}

\bibliographystyle{IEEEtran}
\bibliography{deep}



%\vfill


\end{document}